\documentclass[11pt]{article}
\PassOptionsToPackage{hyperfootnotes=false}{hyperref}
\usepackage[preprint]{acl}

\usepackage{enumitem}
\usepackage{times}
\usepackage{latexsym}
\usepackage[T1]{fontenc}
\usepackage[utf8]{inputenc}
\usepackage{microtype}
\usepackage{inconsolata}
\usepackage{listings}
\usepackage{textcomp}
\usepackage{graphicx}
\usepackage{amsmath}
\usepackage{amssymb}
\usepackage{booktabs}
\usepackage{multirow}
\usepackage{placeins}
\usepackage{tikz}
\usepackage{pgfplots}
\pgfplotsset{compat=1.18}
\usepgfplotslibrary{groupplots}
\usetikzlibrary{arrows.meta,calc,positioning}
\newcommand{\system}{ProactMem}
\title{Proactive Memory for Event-Driven LLM Agents}
\author{
  Jinyang Li$^{1}$ \quad Zichuan Fu$^{1}$ \quad Xian Wu$^{2}$ \\
  \textbf{Wenlin Zhang$^{1}$ \quad Yimin Deng$^{1}$ \quad Kaifeng Guo$^{1}$} \\
  \textbf{Guojing Li$^{1}$ \quad Junjia QI$^{1}$ \quad Xiangyu Zhao$^{1}$}\thanks{Corresponding author.} \\
  $^{1}$City University of Hong Kong \quad $^{2}$Tencent \\
  \texttt{\{jinyali2-c,zc.fu\}@my.cityu.edu.hk} \\
  \texttt{xianzhao@cityu.edu.hk}
}
\hypersetup{
  pdftitle={Proactive Memory for Event-Driven LLM Agents},
  pdfauthor={Jinyang Li, Zichuan Fu, Xian Wu, Wenlin Zhang, Yimin Deng, Kaifeng Guo, Guojing Li, Junjia QI, Xiangyu Zhao}
}

\begin{document}
\maketitle

\begin{abstract}
Proactive LLM agents anticipate service opportunities from ongoing observations and accumulated history.
Existing approaches emphasize recent context or retrieval of related memories, providing limited support for validating recurring behavioral regularities and assessing their applicability to the current event.
We propose Proactive Memory (\system{}), which organizes history into recent events, temporal trajectories, and behavioral regularities.
Trajectory facts are checked against source events, and regularities are assessed through historical replay before being matched to current conditions.
We also construct ProactiveStream, a longitudinal benchmark covering three application scopes, where agents process isolated event streams chronologically under episode-aware opportunity labels.
On ProactiveStream, \system{} achieves an overall effectiveness (F1) of 52.6\%, a 21.4\% relative improvement over the strongest of five external baselines.
Code is available at: \href{https://anonymous.4open.science/r/ProactMem-5351/}{https://anonymous.4open.science/r/ProactMem-5351/}.
\end{abstract}

\section{Introduction}
\label{sec:introduction}

LLM agents are extending task execution toward proactive assistance that identifies service opportunities from ongoing observations and accumulated history.
In activity support, smart homes, and software collaboration, useful assistance depends on recognizing new needs, selecting relevant historical evidence, and tracking prior assistance.

Existing research addresses proactive assistance through service decisions and memory construction.
ProactiveAgent learns task proposals from activity contexts~\citep{lu2025proactiveagent}; ContextAgent combines sensed context with historical personas~\citep{yang2025contextagent}.
Mem0 consolidates and retrieves information~\citep{chhikara2025mem0}, while A-MEM links historical notes~\citep{xu2025amem}.
ProactiveBench and ContextAgentBench evaluate task proposals and contextual service; LoCoMo and LongMemEval assess long-term conversational memory~\citep{maharana2024locomo,wu2025longmemeval}.
Continuous assistance also requires assessing memory accumulation, current applicability, and changing opportunities within one event stream.
This raises a central question: \emph{how can an agent turn accumulated history into reliable evidence for deciding when to offer assistance?}

\begin{figure}[t]
    \centering
    \includegraphics[width=0.96\columnwidth]{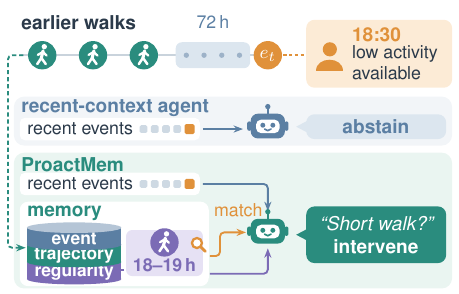}
    \caption{Illustrative Sports scenario: \system{} uses a supported evening-walk regularity beyond recent events.}
    \label{fig:intro_scenario}
\end{figure}

To study this question, we construct \emph{ProactiveStream}, a longitudinal event-stream benchmark spanning Sports, Home, and Code.
Agents build memory chronologically within isolated participant, home, and repository streams, assessing each opportunity's first actionable window and subsequent changes.
With a shared backbone and intervention control policy, existing methods show different trade-offs between opportunity recovery and false interventions (Table~\ref{tab:main_results}).
Figure~\ref{fig:intro_scenario} links earlier evening walks to current assistance.

We propose Proactive Memory (\system{}; Figure~\ref{fig:architecture}), linking recent observations, temporal activities, and recurring behavior.
Event memory preserves recent evidence; temporal trajectory memory abstracts source-grounded activities; behavioral regularity memory captures supported conditional behavior.

LLMs construct cited trajectories from completed events, followed by source validation of typed facts.
Historical replay assesses regularities inferred from these trajectories.
At decision time, contextual assessment combines the current event, recent history, trajectories, and applicable regularities to judge service value.
Intervention control filters repeated proposals; the finalized decision precedes event commitment and memory updates.

Our contributions are:
\begin{itemize}[leftmargin=*, itemsep=0pt]
    \item We introduce ProactiveStream, a longitudinal benchmark for proactive assistance across Sports, Home, and Code, preserving causal history, entity isolation, and episode-aware opportunity labels.
    \item We propose \system{}, a hierarchical memory framework that integrates recent events, source-validated temporal trajectories, and replay-assessed behavioral regularities, matching historical evidence to current conditions.
    \item We find that validated hierarchical memory recovers more service opportunities with fewer false interventions than recent-event memory alone across three scopes.
\end{itemize}

\section{Problem formulation}
\label{sec:problem}

Each participant, household, or repository $u$ has an isolated public event stream
\begin{equation}
E^u=(e^u_1,\ldots,e^u_{T_u}), \qquad
e^u_t=(\tau^u_t,x^u_t),
\label{eq:event_stream}
\end{equation}
where $T_u$ is the stream length and $x^u_t$ contains public fields available at time $\tau^u_t$.
Events follow chronological order with deterministic tie-breaking; we omit $u$ when clear.
State $Z^u_t$ comprises prior observations and finalized outputs for the same entity, together with the resulting memory and intervention control states.
The policy produces
\begin{equation}
(\hat y_t,\hat m_t)=\pi(e^u_t,Z^u_t),
\qquad \hat y_t\in\{0,1\},
\label{eq:generic_decision}
\end{equation}
where $\hat y_t=1$ denotes \emph{intervene}, $\hat y_t=0$ denotes \emph{abstain}, and $\hat m_t$ is an optional evidence-grounded service message.
After finalization, update function $U$ commits the event and output:
\begin{equation}
Z^u_{t+1}=U(Z^u_t,e^u_t,\hat y_t,\hat m_t).
\label{eq:causal_update}
\end{equation}
The objective is to detect timely, actionable \emph{service opportunities} supported by current or prior observations of the same entity.
Evaluation uses fixed reference decisions $y_t\in\{0,1\}$ under Section~\ref{sec:reference_labels}; labels and annotation metadata are reserved for evaluation.

\section{Proactive Memory}
\label{sec:framework}

\begin{figure*}[t]
    \centering
    \makebox[\textwidth][c]{\resizebox{0.98\textwidth}{!}{\input{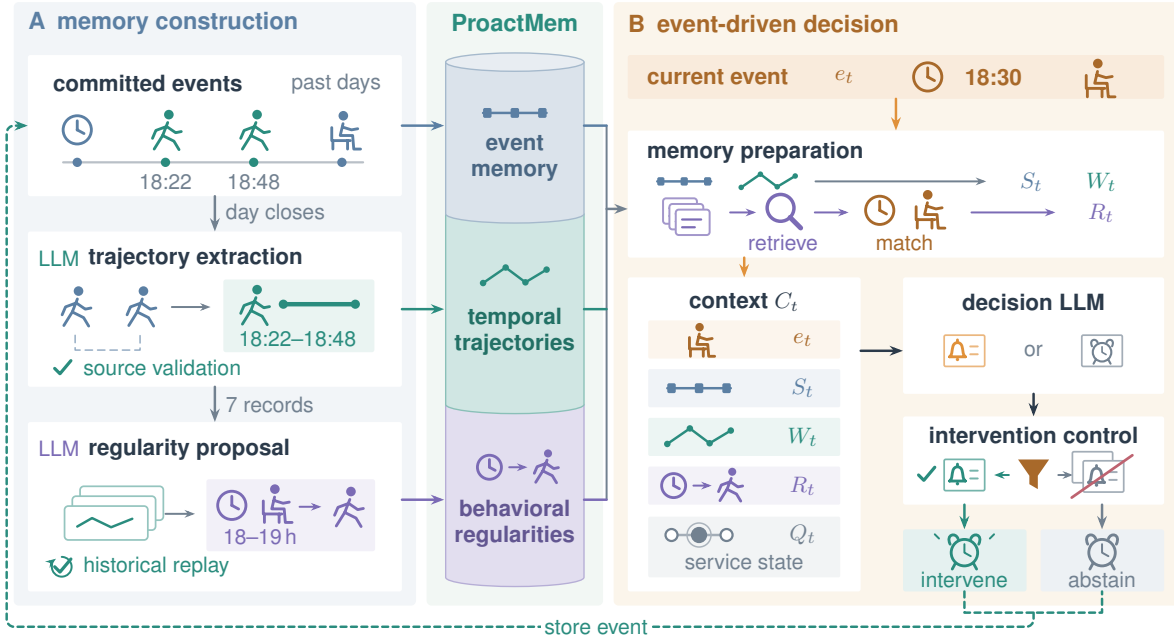}}}
    \caption{\system{} architecture. (A) Source validation and replay build memory. (B) Retrieved memory and current conditions inform assessment and intervention control. Events are stored after the decision (dashed arrows).}
    \label{fig:architecture}
\end{figure*}

\system{} uses three complementary memory layers (Figure~\ref{fig:architecture}): event memory for recent public events, temporal trajectory memory for source-checked activity summaries, and behavioral regularity memory for replay-assessed conditional regularities.
Construction checks source fidelity and historical support.
The decision stage combines current observations with applicable historical evidence through contextual assessment and intervention control.

\subsection{Memory construction and update}

Recent events capture immediate changes; recurring behavior emerges across days.
\system{} organizes these observations into three memory layers.

\paragraph{Bounded event memory.}
Evidence utilization can degrade in long contexts~\citep{liu2024lost}, motivating a bounded event layer that preserves recent observations at their original granularity.
For event $e_t$, event memory retains prior events in their original public representation, intersecting a temporal horizon $\Delta$ with capacity $K$:
\begin{equation}
S_t=\operatorname{Tail}_{K-1}
\left((e_i)_{i<t,\;0\leq\tau_t-\tau_i\leq\Delta}\right).
\label{eq:source_memory}
\end{equation}
Here $\operatorname{Tail}_{K-1}$ selects the most recent $K-1$ events in the filtered sequence.
We use $\Delta=72$ hours and $K=100$: at most 99 prior events accompany the separately supplied current event.
A separate buffer retains all committed events from the open local day for trajectory construction.

\paragraph{Fact-validated temporal trajectory memory.}
At the end of each local day, an LLM summarizes that day's events into trajectory segments with intervals, typed facts, and source citations.
Let $F(\cdot)$ denote normalized checkable facts obtained from typed fields.
For segment $d$ and its cited events $\mathcal{E}(d)$, source consistency requires
\begin{equation}
\begin{aligned}
F(d)&\subseteq\bigcup_{e\in\mathcal{E}(d)}F(e),\\
\mathcal{E}(d)&\subseteq E^u_{\text{completed day}}.
\end{aligned}
\label{eq:daily_validation}
\end{equation}
Here $E^u_{\text{completed day}}$ contains the same entity's events from that completed day.
The validator checks entity, interval, citations, and canonicalized scalar values at compatible field paths.
Acceptance requires source-supported typed facts and resolved citations.
Source citations preserve the link between an accepted trajectory and the observations it summarizes.
An unsupported detail could otherwise propagate into a proposed regularity as daily records are consolidated.
Validation therefore constrains the factual basis on which regularities are proposed, while retaining the source links needed to inspect their support.
Accepted temporal trajectories form $W_t$, retaining historical context beyond event memory.
Appendix~\ref{app:implementation} details field handling and buffering.

\paragraph{Replay-assessed behavioral regularity memory.}
Historical replay tests whether a proposed behavioral regularity recurs in the recorded observations.
Weekly consolidation groups seven completed daily records per entity in time order, with each record assigned to one cycle.
A cycle may span more than seven calendar days.
An LLM proposes a trigger $g$, expected behavior $p$, time window, optional weekday restrictions, and source daily identifiers.
Replay checks these conditions against validated daily facts: an observable triggered day is fulfilled when $p$ occurs at or after the trigger within the window and violated otherwise.
Missing observations are censored, inapplicable days are excluded, and each date contributes at most once.
Let $n_f$ count fulfilled (supporting) days and $n_v$ count violated days:
\begin{equation}
\begin{aligned}
\operatorname{conf}_{\mathrm{replay}}&=\frac{n_f}{n_f+n_v},\\
\operatorname{lift}&=\frac{\operatorname{conf}_{\mathrm{replay}}}{\operatorname{base}(p)}.
\end{aligned}
\label{eq:replay_gate}
\end{equation}
The base rate is the frequency of $p$ over observable days under the same time-window and weekday restrictions across both triggered and other observable days.
A zero denominator gives a zero score.
Confidence measures consistency across observable triggered days, whereas lift accounts for how common the expected behavior is under comparable temporal restrictions.
A frequently observed activity can yield high confidence even when it also occurs without the trigger.
Comparing the two frequencies makes this distinction explicit when assessing a proposed regularity.
The support requirement further ensures that activation rests on repeated observations across distinct days.

A well-formed proposal remains a \emph{candidate} until $n_f\geq3$, confidence $\geq0.5$, and lift $\geq1.0$; it then becomes \emph{active} and available for retrieval.
Malformed proposals are rejected.
Consolidation resets the online trajectory view, retaining records for audit.
Later observations update regularity confidence and lifecycle: repeated violations can lower retrieval priority or suspend eligibility.
Appendix~\ref{app:lifecycle} specifies these transitions.

\subsection{Memory retrieval and selection}

The event memory supplies the bounded sequence and $W_t$ supplies all completed temporal trajectories in the current cycle.
Regularity retrieval checks trigger context for the same entity, weekdays, and temporal applicability with a 30-minute lead and grace interval.
Matching considers current and previously recorded causal trigger occurrences.
An occurrence is fulfilled when its expected behavior has been observed; censored occurrences have insufficient observations for assessment.
Selection excludes both states.
For regularity store $L_t^u$, lifecycle eligibility $\operatorname{Eligible}_t$, trigger and time matching $\operatorname{Match}_t$, and occurrence state $s_t$,
\begin{equation}
\begin{aligned}
R_t&=\operatorname{Top3}\big\{\ell\in L_t^u:\\
&\quad\operatorname{Eligible}_t(\ell)\land\operatorname{Match}_t(\ell),\\
&\quad s_t(\ell)\notin\{\text{fulfilled},\text{censored}\}\big\}.
\end{aligned}
\label{eq:long_term_retrieval}
\end{equation}
The top-three selection ranks eligible regularities by occurrence state, lifecycle, confidence, and temporal distance (Appendix~\ref{app:matching}).
Consider the evening-walk example in Figure~\ref{fig:intro_scenario}.
Repeated walks under similar conditions may support a regularity even after their source events leave recent memory.
When a later event satisfies its trigger near the usual walking time, the regularity contributes evidence from those earlier days.
A walk already completed during the applicable occurrence removes it from selection, as does an occurrence \mbox{with insufficient observations}.

\subsection{Proactive decision}
\label{sec:decision_control}

\paragraph{Contextual assessment and intervention control.}
For each event, the decision context combines the current event, recent events, temporal trajectories, selected regularities, and observable episode state:
\begin{equation}
C_t=(e_t,S_t,W_t,R_t,Q_t),
\label{eq:decision_context}
\end{equation}
where $Q_t$ describes the service opportunity's observable state; the controller separately records previously allowed interventions to track service coverage.
The LLM produces $(\tilde y_t,\tilde m_t)=\pi_\theta(C_t)$, proposing intervention when the current event \emph{or} historical context supports concrete, timely, feasible, low-risk assistance, and abstention otherwise.
For the walking example, a retrieved regularity supplies the historical reason to consider a reminder.
The assessment then uses recent activity and the current situation to determine whether such assistance is timely and feasible.

Within each scope, all methods apply the same intervention controller policy to their own context and intervention history.
The controller compares proposed services with previously allowed interventions and converts duplicates to \emph{abstain}.
Scope-specific changes in opportunity state, service identity, contextual regularities, or local day permit renewed assessment; each intervention originates from an LLM proposal.
The controller finalizes the proposal as $(\hat y_t,\hat m_t)$, the output defined in Section~\ref{sec:problem}.
Appendix~\ref{app:intervention_control} specifies matching fields, phase transitions, and re-entry rules.

Inference stages $e_t$, retrieves prior memory and matches current conditions, assembles $C_t$ for assessment, applies intervention control, and commits $e_t$ after finalization.
Each complete public event enters the event buffer and open-day buffer, advancing trajectory construction, regularity replay, and lifecycle updates.
Abstained events can still record completed activities, changed circumstances, or violations of an existing regularity.
Storing them preserves the observations needed to revise memory as behavior changes.
Maintenance incorporating $e_t$ can affect only later decisions.
The current event is therefore assessed against the memory available on arrival, together with its own observed content.
Reference labels are reserved for offline evaluation.

\begin{table*}[t]
\centering
\caption{Composition and provenance of ProactiveStream. Reference interventions mark service opportunities.}
\label{tab:benchmark_stats}
\resizebox{\textwidth}{!}{%
\begin{tabular}{@{}lcccccc@{}}
\toprule
\textbf{Scope} & \textbf{Primary Source} & \textbf{Entity} & \textbf{Streams} & \begin{tabular}[c]{@{}c@{}}\textbf{Input}\\ \textbf{Events}\end{tabular} & \textbf{Temporal Coverage} & \begin{tabular}[c]{@{}c@{}}\textbf{Reference}\\ \textbf{Interventions}\end{tabular} \\ \midrule
Sports & HeartSteps     & participant     & 9       & 1,456                                                  & 367 observed participant-days     & 365                                                               \\
Home   & CASAS          & home            & 4       & 1,624                                                  & 14 consecutive days/home          & 190                                                               \\
Code   & GH Archive     & repository      & 4       & 1,459                                                  & longitudinal repository histories & 174                                                               \\\midrule
Total  & three scopes   & isolated stream & 17      & 4,539                                                  & three longitudinal settings       & 729                                                               \\ \bottomrule
\end{tabular}%
}
\end{table*}

\section{ProactiveStream benchmark}
\label{sec:benchmark}

ProactiveStream evaluates service-opportunity detection in longitudinal streams as agents accumulate history.
Table~\ref{tab:benchmark_stats} gives its composition.

\subsection{Scopes and public events}

\hspace*{\parindent}\textbf{Sports} uses participant activity and availability contexts for optional walking and everyday physical-activity assistance.

\textbf{Home} uses time-bounded home-sensor observations for gentle, low-risk check-ins.
Each sensor segment becomes available at interval closure.

\textbf{Code} uses repository activity at GitHub's native event granularity for low-risk triage, review, and maintenance, with one decision per event.

Each participant, home, or repository defines an isolated chronological stream, allowing earlier observations to inform later service opportunities.
Public inputs contain contemporaneously available observations, using the same scope-specific schema for both label classes.
Agents construct their own memory from these events; annotation and scoring metadata are reserved for evaluation.
Appendix~\ref{app:benchmark_protocol} details source datasets, fields, and preprocessing.

\subsection{Reference labels and causal replay}
\label{sec:reference_labels}

Annotators assess $e_t$ and its complete prior history $\mathcal{H}_{<t}$ within the same stream.
Let $A$ indicate a concrete, timely, feasible, low-risk opportunity supported by $e_t$, $B$ an opportunity at $e_t$ supported by historical observations, and $N_t^{\mathrm{ref}}$ the novelty of the reference episode, shared across methods:
\begin{equation}
\begin{aligned}
y_t=1 \iff {} & \big(A(e_t)=1\\
& {}\lor B(e_t,\mathcal{H}_{<t})=1\big)\\
& {}\land N_t^{\mathrm{ref}}=1.
\end{aligned}
\label{eq:or_semantics}
\end{equation}
An event can provide enough evidence for immediate assistance, as represented by $A$.
For $B$, its significance depends on earlier observations: a recurring activity or an earlier unresolved need can make a seemingly routine event actionable.
The complete prior history available to annotators allows such opportunities to be recognized even when the supporting events fall outside a method's bounded context.
$A$ and $B$ may both hold for the same event; each requires a concrete service opportunity at the current time.

The reference protocol labels the first reliable actionable window \emph{intervene} and covered windows \emph{abstain}.
This assigns a detection point to an opportunity that may span several events.
Later events can reopen assessment when support strengthens, the service changes, or an episode reopens.
Event-level labels therefore capture when useful assistance becomes warranted, together with the persistence and renewal of the underlying opportunity.

LLM-assisted triage prioritizes candidate opportunities; three team members then independently label every event, followed by cross-review and adjudication.
Appendix~\ref{app:label_review_450} details a separate-team diagnostic of reference sensitivity and episode behavior.
Main comparisons fix labels, prompts, memory bounds, and validation settings before evaluation.

Each stream is replayed chronologically from empty method state, using the current event and strictly prior context from that stream.
Each decision is finalized before the current event is committed to memory; reference labels are joined after inference for scoring.
Scoring includes the initial events and continues as history accumulates.

\begin{table*}[t]
\centering
\caption{Full-scope comparison (GPT-5.6 Luna, \%; frequency in $\times$). Bold/underline: best/second-best per scope.}
\label{tab:main_results}
\resizebox{\textwidth}{!}{%
\begin{tabular}{@{}clcccccc@{}}
\toprule
\textbf{Scope} & \textbf{Method} & \textbf{Frequency} ($\times$) & \textbf{Relevance} $\uparrow$ & \textbf{Coverage} $\uparrow$ & \textbf{Effectiveness} $\uparrow$ & \textbf{Intrusion} $\downarrow$ & \textbf{Balance} $\uparrow$ \\
\midrule
\multirow{6}{*}{Sports} & ProactiveAgent & 2.75 & 30.0 & \underline{82.5} & 43.9 & 64.5 & \underline{59.0} \\
  & ContextAgent & 2.16 & \underline{31.0} & 67.1 & 42.4 & \underline{50.0} & 58.6 \\
  & Mem0 & 3.40 & 28.8 & \textbf{97.8} & \underline{44.5} & 81.0 & 58.4 \\
  & A-MEM & 2.41 & 30.6 & 74.0 & 43.3 & 56.0 & \underline{59.0} \\
  & Graphiti & 2.20 & 29.3 & 64.4 & 40.3 & 52.0 & 56.2 \\
  & \system{} & 2.15 & \textbf{35.8} & 76.7 & \textbf{48.8} & \textbf{46.1} & \textbf{65.3} \\
\midrule
\multirow{6}{*}{Home} & ProactiveAgent & 0.78 & 34.2 & 26.8 & 30.1 & \textbf{6.8} & 60.0 \\
  & ContextAgent & 1.38 & \underline{35.9} & 49.5 & 41.6 & 11.7 & 68.9 \\
  & Mem0 & 0.73 & 23.9 & 17.4 & 20.1 & \underline{7.3} & 55.0 \\
  & A-MEM & 1.63 & 34.6 & \underline{56.3} & \underline{42.9} & 14.1 & \underline{71.1} \\
  & Graphiti & 1.53 & 33.4 & 51.1 & 40.4 & 13.5 & 68.8 \\
  & \system{} & 1.52 & \textbf{48.6} & \textbf{73.7} & \textbf{58.6} & 10.3 & \textbf{81.7} \\
\midrule
\multirow{6}{*}{Code} & ProactiveAgent & 4.45 & 17.5 & \underline{78.2} & 28.7 & 49.7 & 64.2 \\
  & ContextAgent & 2.68 & 22.5 & 60.3 & 32.8 & 28.2 & 66.1 \\
  & Mem0 & 0.65 & \underline{39.8} & 25.9 & 31.4 & \textbf{5.3} & 60.3 \\
  & A-MEM & 2.55 & 30.4 & 77.6 & \underline{43.7} & 24.0 & \underline{76.8} \\
  & Graphiti & 2.36 & 26.8 & 63.2 & 37.7 & 23.3 & 69.9 \\
  & \system{} & 2.24 & \textbf{40.1} & \textbf{89.7} & \textbf{55.4} & \underline{18.1} & \textbf{85.8} \\
\bottomrule
\end{tabular}%
}
\end{table*}

\section{Experimental evaluation}
\label{sec:experiments}

We examine overall effectiveness (\textbf{RQ1}), model choice (\textbf{RQ2}), and memory organization and validation (\textbf{RQ3}), with supplementary checks of reference sensitivity and episode behavior.

\subsection{Experimental settings}
\label{sec:experimental_settings}

\begin{table*}[t]
\centering
\caption{Within-series model scaling (\%). Bold and underline mark the best and second-best scores per column.}
\label{tab:backbone_sensitivity}
\resizebox{\textwidth}{!}{%
\begin{tabular}{@{}lcccccc@{}}
\toprule
\multirow{2}{*}{\textbf{Model}} & \multicolumn{2}{c}{\textbf{Sports}} & \multicolumn{2}{c}{\textbf{Home}} & \multicolumn{2}{c}{\textbf{Code}} \\
\cmidrule(lr){2-3}\cmidrule(lr){4-5}\cmidrule(lr){6-7}
 & \textbf{Effectiveness} $\uparrow$ & \textbf{Intrusion} $\downarrow$ & \textbf{Effectiveness} $\uparrow$ & \textbf{Intrusion} $\downarrow$ & \textbf{Effectiveness} $\uparrow$ & \textbf{Intrusion} $\downarrow$ \\
\midrule
GPT-5.6 Luna & 48.8 & 46.1 & 58.6 & 10.3 & 55.4 & 18.1 \\
GPT-5.6 Terra & \underline{54.1} & \underline{39.4} & \underline{64.4} & \underline{9.1} & \underline{60.3} & \underline{14.9} \\
GPT-5.6 Sol & \textbf{58.6} & \textbf{34.4} & \textbf{69.3} & \textbf{7.5} & \textbf{66.4} & \textbf{11.8} \\
\bottomrule
\end{tabular}%
}
\end{table*}

\begin{figure*}[tp]
\centering
\includegraphics[width=0.32\textwidth]{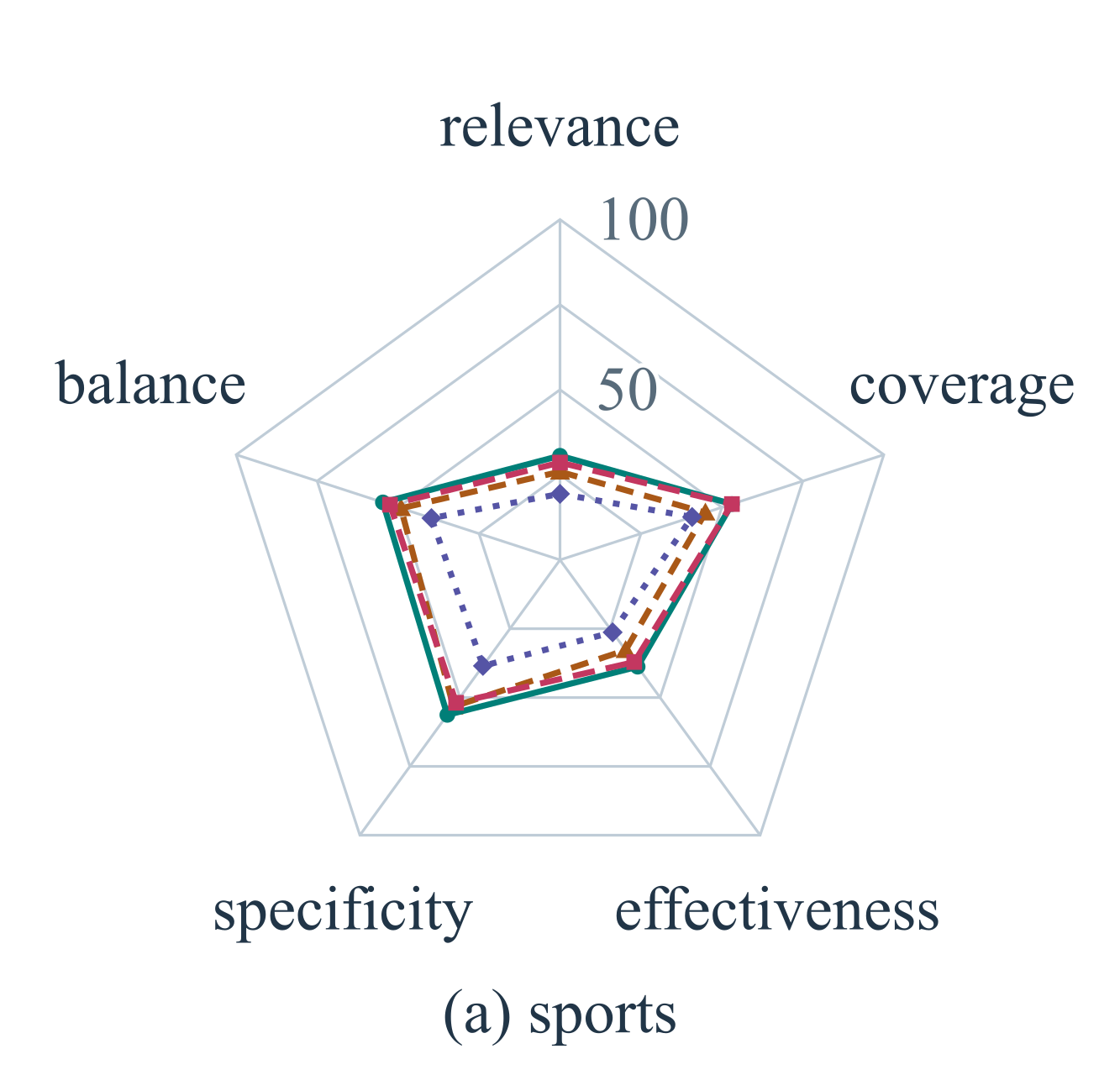}\hfill
\includegraphics[width=0.32\textwidth]{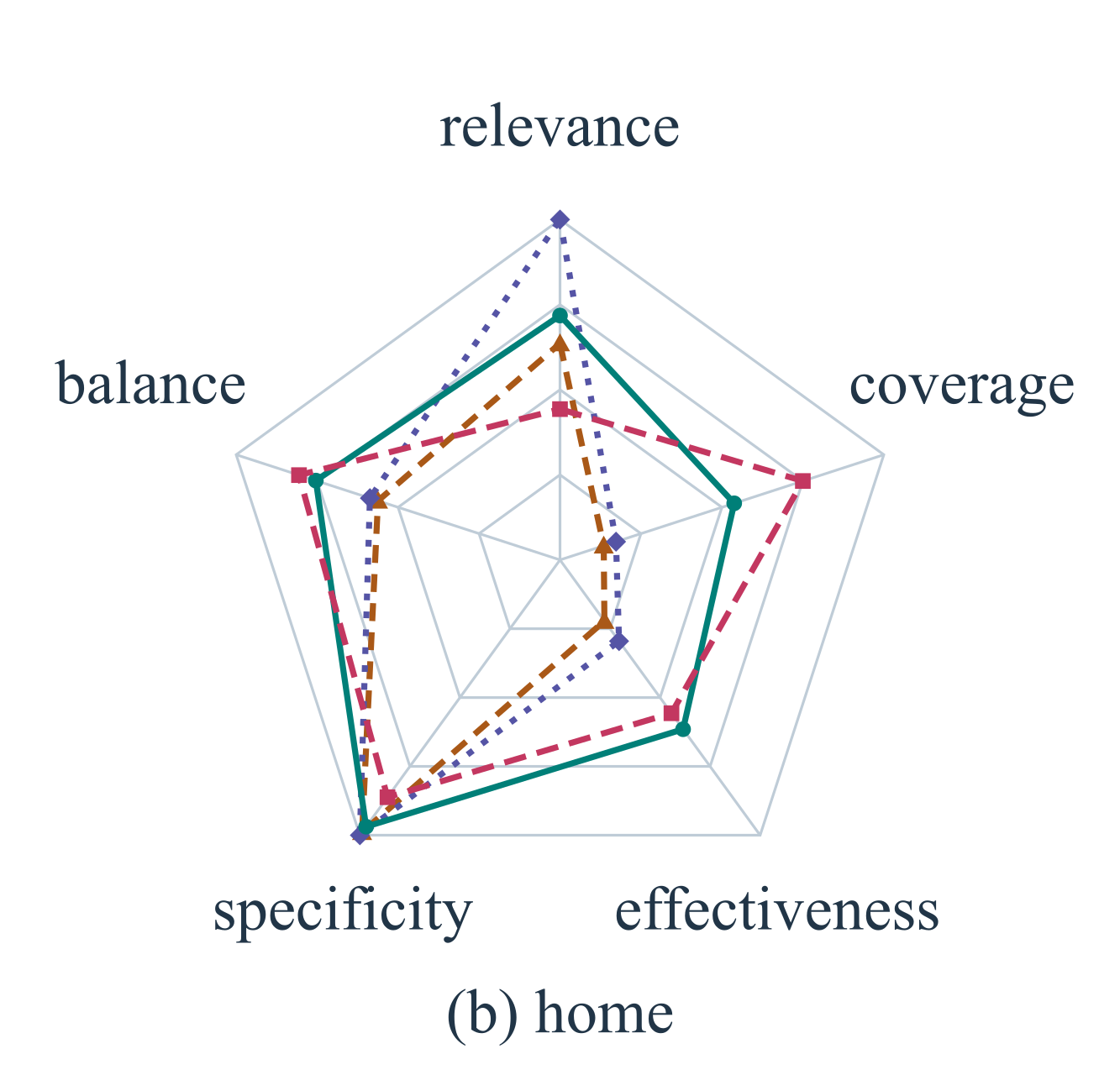}\hfill
\includegraphics[width=0.32\textwidth]{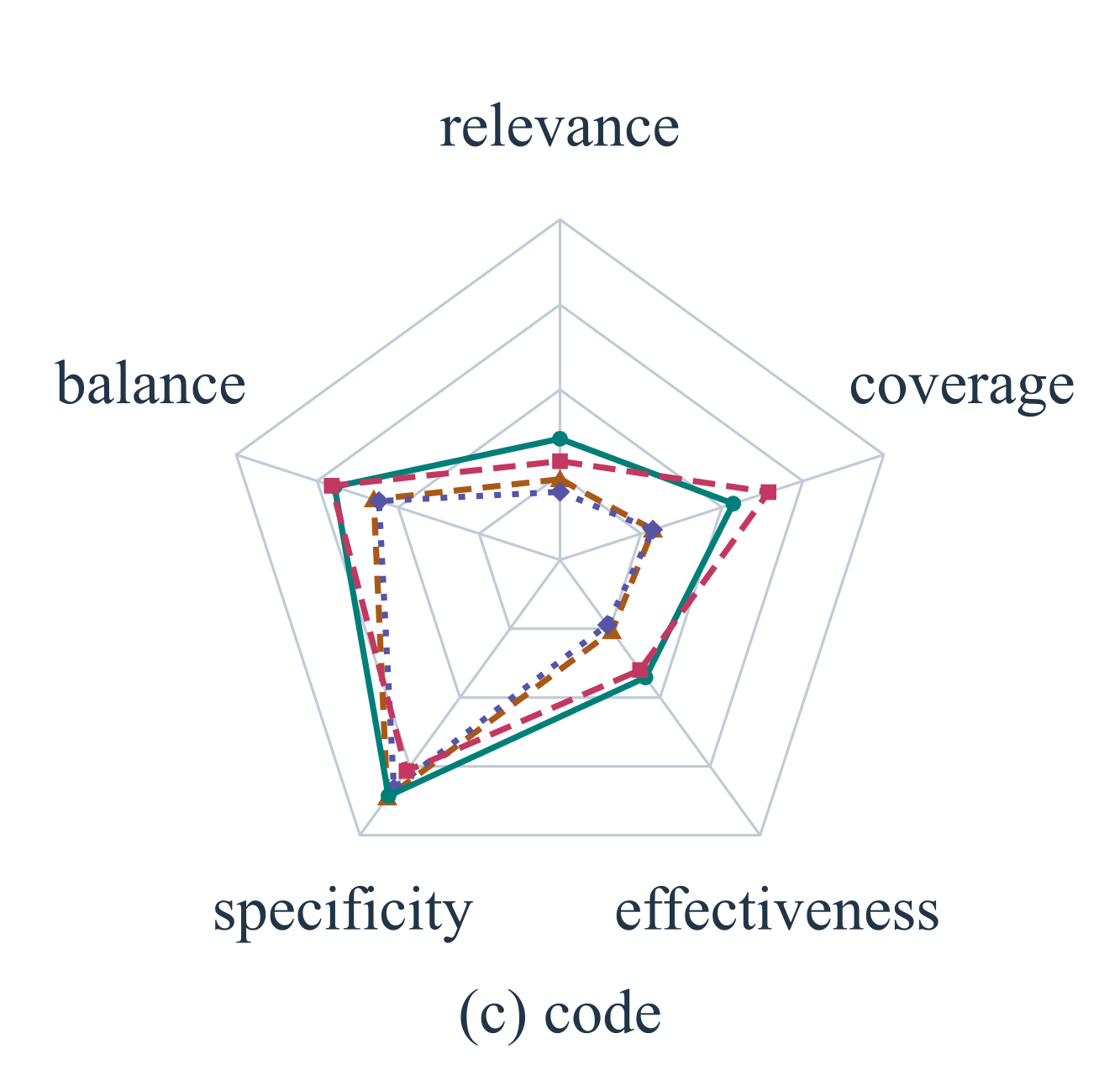}
\par\vspace{2pt}
\includegraphics[width=0.98\textwidth]{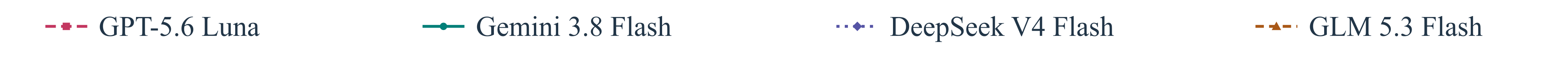}
\caption{Event-only comparison of four backbones on selected streams (0--100\%). Specificity $=1-\mathrm{intrusion}$.}
\label{fig:event_only_model_radar}
\end{figure*}

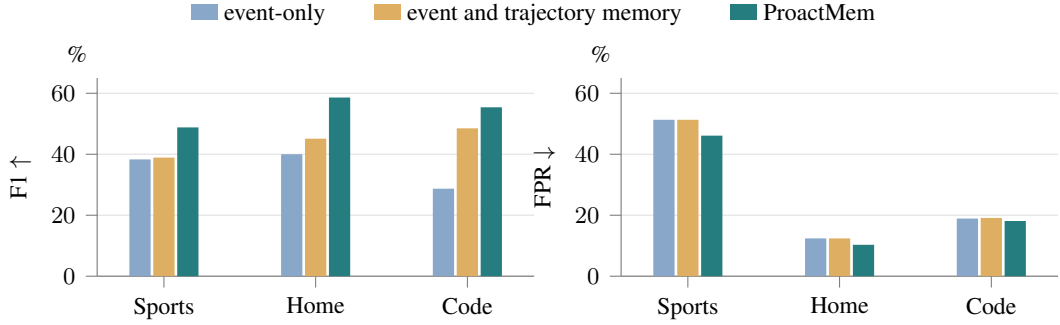
\begin{figure*}[t]
\centering
\definecolor{eventblue}{HTML}{89A7C8}
\definecolor{dailygold}{HTML}{DEAF62}
\definecolor{patternteal}{HTML}{267D80}
{\small
\textcolor{eventblue}{\rule{10pt}{6pt}}\;event-only\hspace{16pt}
\textcolor{dailygold}{\rule{10pt}{6pt}}\;event and trajectory memory\hspace{16pt}
\textcolor{patternteal}{\rule{10pt}{6pt}}\;\system{}\par}
\smallskip
\begin{tikzpicture}
\pgfplotsset{memorybars/.style={
    ybar=1pt,bar width=8pt,width=0.46\textwidth,height=4.2cm,
    symbolic x coords={Sports,Home,Code},xtick=data,enlarge x limits=0.22,
    ymin=0,ymax=65,ytick={0,20,40,60},clip=false,
    axis x line*=bottom,axis y line*=left,tick align=outside,
    axis line style={black!55},tick style={black!55},
    ymajorgrids=true,grid style={black!12},
    font=\small,ylabel style={font=\small},
    every axis plot/.append style={draw=none}
}}
\begin{axis}[memorybars,name=ispanel,ylabel={F1 $\uparrow$}]
\node[anchor=south east,font=\small,yshift=2pt] at (rel axis cs:0,1) {\%};
\addplot[fill=eventblue] coordinates {(Sports,38.3) (Home,40.0) (Code,28.7)};
\addplot[fill=dailygold] coordinates {(Sports,38.9) (Home,45.1) (Code,48.5)};
\addplot[fill=patternteal] coordinates {(Sports,48.8) (Home,58.6) (Code,55.4)};
\end{axis}
\begin{axis}[memorybars,at={(ispanel.east)},anchor=west,xshift=1.15cm,ylabel={FPR $\downarrow$}]
\node[anchor=south east,font=\small,yshift=2pt] at (rel axis cs:0,1) {\%};
\addplot[fill=eventblue] coordinates {(Sports,51.3) (Home,12.4) (Code,18.9)};
\addplot[fill=dailygold] coordinates {(Sports,51.3) (Home,12.4) (Code,19.1)};
\addplot[fill=patternteal] coordinates {(Sports,46.1) (Home,10.3) (Code,18.1)};
\end{axis}
\end{tikzpicture}
\caption{Memory-layer ablation (GPT-5.6 Luna) under a shared decision prompt and intervention control policy.}
\label{fig:memory_hierarchy}
\end{figure*}

\begin{table*}[t]
\centering
\caption{Validation and Code intervention control (GPT-5.6 Luna, \%). Each w/o row drops one check; before suppression scores recorded proposals with later states fixed. Bold and underline mark the best and second-best.}
\label{tab:validation_ablations}
\resizebox{\textwidth}{!}{%
\begin{tabular}{lcccccc}
\toprule
& \multicolumn{2}{c}{\textbf{Sports}} & \multicolumn{2}{c}{\textbf{Home}} & \multicolumn{2}{c}{\textbf{Code}} \\
\cmidrule(lr){2-3}\cmidrule(lr){4-5}\cmidrule(lr){6-7}
\textbf{Method} & \textbf{Relevance} & \textbf{Effectiveness} & \textbf{Relevance} & \textbf{Effectiveness} & \textbf{Relevance} & \textbf{Effectiveness} \\
\midrule
w/o replay gate & 31.1 & 42.7 & 43.6 & 54.4 & 33.8 & 48.1 \\
w/o trajectory validation & \underline{34.4} & \underline{47.5} & \underline{45.9} & \underline{56.7} & 36.1 & 51.5 \\
before suppression & -- & -- & -- & -- & \underline{36.7} & \underline{53.6} \\
\system{} & \textbf{35.8} & \textbf{48.8} & \textbf{48.6} & \textbf{58.6} & \textbf{40.1} & \textbf{55.4} \\
\bottomrule
\end{tabular}%
}
\end{table*}

\paragraph{Baselines.}
We evaluate five external methods:
\begin{list}{\labelitemi}{%
    \settowidth{\labelwidth}{\labelitemi}%
    \setlength{\labelsep}{0.5em}%
    \setlength{\leftmargin}{\labelwidth}%
    \addtolength{\leftmargin}{\labelsep}%
    \setlength{\topsep}{0pt}%
    \setlength{\partopsep}{0pt}%
    \setlength{\itemsep}{0pt}%
    \setlength{\parsep}{0pt}%
}
    \item \textbf{ProactiveAgent} uses its released prompt and proposal-or-null output contract, with causal history of events and decisions capped at 1,000 events~\citep{lu2025proactiveagent,thunlp2026proactiveagentrepo}.
    \item \textbf{ContextAgent} adapts proactive-service judgment to public events~\citep{yang2025contextagent}.
    \item \textbf{Mem0} consolidates and retrieves memories for the same entity~\citep{chhikara2025mem0,mem0ai2026mem0v2012}.
    \item \textbf{A-MEM} updates linked notes and associations to retrieve relevant notes~\citep{xu2025amem}.
    \item \textbf{Graphiti} stores temporal event relations and retrieves facts from the same stream~\citep{zep2026graphiti}.
\end{list}

\paragraph{Metrics.}
Each event receives one decision finalized by intervention control; immediate and scheduled services count as \emph{intervene} at the decision event.
We report \textbf{relevance} (precision), the share of interventions matching reference opportunities; \textbf{coverage} (recall), the share of opportunities recovered; and \textbf{effectiveness} (F1), their harmonic mean.
Overall effectiveness pools confusion counts across all streams and scopes.
The \textbf{intrusion} score (false positive rate) measures the share of reference negatives receiving interventions; \textbf{balance} (balanced accuracy) averages coverage and the correct-abstention rate on reference negatives.
Higher relevance, coverage, effectiveness, and balance and lower intrusion are preferred.
The \textbf{frequency} metric measures interventions per reference-positive event: $(\mathrm{TP}+\mathrm{FP})/(\mathrm{TP}+\mathrm{FN})$, with counts pooled across streams within each scope.
Appendix~\ref{app:configuration} gives metric formulas and frequency interpretation.

\paragraph{Implementation details.}
The primary model is \texttt{gpt-5.6-luna}, with temperature zero and thinking disabled; \system{} uses the official OpenAI API.
Methods share task semantics and the intervention controller policy, with interface-specific prompts and budgets.
The event-only, event and trajectory memory, and \system{} variants share their decision prompt, schema, and context bound of 72 hours and at most 100 events, including the current event.
A-MEM and Graphiti augment that bound with their representations.
For \system{}, each event undergoes contextual assessment with prepared memory and current-condition matches, followed by intervention control.
Appendix~\ref{app:configuration} lists detailed configurations.

\subsection{Main results (RQ1)}
\label{sec:main_results}

We compare \system{} with five external methods on ProactiveStream.

\paragraph{Comparisons with external methods.}
Across all three scopes, \system{} achieves higher effectiveness and balance than the external baselines (Table~\ref{tab:main_results}, RQ1).
The comparison reflects different trade-offs across scopes.
On Sports, Mem0 attains greater coverage alongside more false interventions; \system{} is more selective, with higher relevance and lower intrusion.
On Home, \system{} recovers more opportunities than every external baseline, with fewer false interventions than the higher-frequency A-MEM and Graphiti.
On Code, it likewise obtains the highest coverage, whereas Mem0 achieves lower intrusion with substantially less opportunity recovery.
The gains therefore come with different changes in operating behavior: more selective intervention on Sports and broader opportunity recovery on Home and Code.

\label{sec:reviewed_diagnostics}
Appendix~\ref{app:label_review_450} examines reference sensitivity and episode behavior on a separate diagnostic sample.
Rescoring fixed diagnostic outputs against separate-team references gives similar effectiveness on Sports and lower effectiveness on Home and Code, indicating that absolute performance depends on how actionable opportunities are labeled.

\subsection{Model comparison (RQ2)}
\label{sec:model_sensitivity}

\paragraph{Within-series model scaling.}
With all other settings fixed, construction and decision LLMs change together (Table~\ref{tab:backbone_sensitivity}).
Effectiveness increases and intrusion decreases from GPT-5.6 Luna through GPT-5.6 Terra to GPT-5.6 Sol across scopes.
Stronger backbones can improve both the construction of conditional regularities and their use in decisions about the current event.
Controlled ablations retain GPT-5.6 Luna.

\paragraph{Cross-family event-only comparison.}
Using the shared event-only interface, we compare GPT-5.6 Luna, Gemini 3.8 Flash, DeepSeek V4 Flash, and GLM 5.3 Flash on two selected complete streams per scope (Figure~\ref{fig:event_only_model_radar}; Appendix~\ref{app:cross_family_streams}).
On Home and Code, Luna recovers more opportunities than DeepSeek and GLM, while the latter achieve higher specificity.
Sports provides a contrast: Luna achieves both higher coverage and higher specificity than DeepSeek.

\subsection{Ablation and memory analysis (RQ3)}
\label{sec:hierarchy}
\label{sec:validation_ablations}

\paragraph{Proactive agent ablation.}
Figure~\ref{fig:memory_hierarchy} compares event-only memory, event and trajectory memory, and the full hierarchy using Luna under shared decision settings.
Trajectories extend activity context beyond recent events, while regularities organize recurring evidence around its conditions of applicability.
Adding trajectory memory raises effectiveness in every scope, with the largest improvement on Code.
Intrusion remains unchanged on Sports and Home and rises slightly on Code, so the gain at this stage lies mainly in opportunity recovery.

The full hierarchy improves effectiveness further and lowers intrusion across all three scopes.
It retains recurring behavior together with its triggers and time windows, allowing earlier observations to inform assessment when the corresponding conditions recur.
The improvement on both measures is consistent with a useful role for this more selective historical context.
These configurations also use representation-dependent intervention keys, so the comparison captures memory retrieval and intervention control operating together.

\paragraph{Memory validation.}
Source validation checks trajectory fidelity; replay assesses historical support for conditional regularities.
Table~\ref{tab:validation_ablations} compares both ablations with the same model and decision settings.
\emph{w/o replay gate} retains current matching and bypasses activation thresholds for well-formed, source-linked candidates reconstructed online.
\emph{w/o trajectory validation} retains schema-valid extractions for the same entity, weekly construction, and replay, omitting fact checks.
Both reduce relevance and effectiveness; removing replay yields the larger effectiveness decline.
On Code, trajectory validation is associated with fewer false interventions at unchanged coverage (RQ3).

\paragraph{Intervention control.}
The Code analysis reveals a trade-off between coverage and selectivity.
On recorded proposals, suppression removes more false positives than true positives, improving relevance and effectiveness while reducing coverage.
Later states and decisions stay fixed (Appendix~\ref{app:extended_ablations}).

\section{Related work}
\label{sec:related}

\paragraph{Agent memory.}
Generative Agents, MemoryBank, and MemGPT organize experience through reflection, memory updates, or tiered context management~\citep{park2023generative,zhong2024memorybank,packer2023memgpt}.
For retrieval, Mem0 uses extracted memories, A-MEM links records, and HippoRAG relies on graphs~\citep{chhikara2025mem0,xu2025amem,gutierrez2024hipporag}.
Graphiti also tracks when event-derived relations are valid and where they originate~\citep{zep2026graphiti}.

\paragraph{Proactive agents and memory.}
ProactiveAgent studies whether to propose a task or remain silent given annotated events~\citep{lu2025proactiveagent}.
To anticipate service needs, ContextAgent combines sensory contexts with historical personas~\citep{yang2025contextagent}, and PASK couples streaming demand detection with hybrid long-term memory~\citep{xie2026pask}.
\mbox{MemCog} explores memory for dialogue reasoning~\citep{li2026memcog}; Remember When It Matters uses a separate memory agent to supply selected reminders for long-horizon action~\citep{wu2026remember}.
\system{} checks trajectory facts against source events and tests regularities through replay before matching them to current conditions.

\paragraph{Evaluation of proactive service.}
LoCoMo and LongMemEval assess long-term conversational memory and reasoning~\citep{maharana2024locomo,wu2025longmemeval}.
For proactive agents, benchmarks examine task proposals, latent-need detection, or memory triggering~\citep{lu2025proactiveagent,xie2026pask,li2026memcog}.
ProactiveStream follows successive \emph{intervene} or \emph{abstain} decisions as memory accumulates, using isolated, causally replayed streams and episode-aware references for intervention timing.
Interruption costs vary with task phase and user receptivity~\citep{adamczyk2004ifnotnow,iqbal2007breakpoints,iqbal2008notification,mehrotra2016myphone}.
We therefore evaluate both \mbox{opportunity recovery and intrusion}.

\FloatBarrier
\section{Conclusion}
\label{sec:conclusion}

We introduced ProactiveStream to evaluate proactive assistance over event streams as history accumulates.
\system{} combines recent events, source-checked temporal trajectories, and replay-assessed behavioral regularities, bringing evidence from earlier observations into current decisions.
Across Sports, Home, and Code, it achieves higher effectiveness and balance than the evaluated baselines.
The memory ablations further show improved opportunity recovery with trajectories and fewer false interventions with the full hierarchy.
These results support maintaining historical evidence at multiple temporal scales for \mbox{event-driven proactive assistance}.

\label{sec:main_end}
\section*{Limitations}
\label{sec:limitations}

ProactiveStream covers low-risk service settings in Sports, Home, and Code under a protocol fixed after joint method and benchmark development.
Independent held-out evaluation and cross-family transfer of constructed memory remain future work.
Reference labels represent service opportunities under the annotation protocol; the 450-target review assesses three selected streams.
The Code suppression diagnostic holds later states fixed; controller-free behavior requires a sequential comparison.
Historical replay assesses observed support; prospective validity under changing behavior requires further study.
Our offline metrics evaluate opportunity detection.
Recipient benefit, interruption burden, end-to-end cost, and deployment safety require prospective measurements.
The reported results support comparisons within the evaluated protocols, with broader generalization depending on additional populations, observation conditions, and model families.

\section*{Ethical considerations}

Our evaluation is limited to optional activity support, gentle check-ins, and low-risk repository assistance.
Medical and high-risk safety decisions require separate domain-specific validation and safeguards.
Inference is isolated by stream, and reference labels, episode annotations, and scoring metadata are excluded from model inputs.

Longitudinal events and derived memories can reveal sensitive routines even when identifiers are pseudonymized.
Event content, prediction traces, and inferred trajectories require access and retention protections; external-model use should limit data transmission and respect source permissions.

Deployment should require informed opt-in and allow inspection, correction, deletion, retention control, and disabling proactive assistance.
An intervention prediction represents a service proposal; consequential action requires separate authorization.
These deployment safeguards require separate implementation and evaluation.

\begingroup
\let\paperbibliography\thebibliography
\renewcommand{\thebibliography}[1]{%
  \paperbibliography{#1}%
  \setlength{\rightskip}{0pt plus 2em}}
\bibliography{9Reference}

\begin{thebibliography}{24}
\providecommand{\natexlab}[1]{#1}

\bibitem[{Adamczyk and Bailey(2004)}]{adamczyk2004ifnotnow}
Piotr~D. Adamczyk and Brian~P. Bailey. 2004.
\newblock \href {https://doi.org/10.1145/985692.985727} {If not now, when?: The
  effects of interruption at different moments within task execution}.
\newblock In \emph{Proceedings of the SIGCHI Conference on Human Factors in
  Computing Systems}, \unskip\nobreakspace{}pages~271--278.

\bibitem[{Chhikara et~al.(2025)Chhikara, Khant, Aryan, Singh, and
  Yadav}]{chhikara2025mem0}
Prateek Chhikara, Dev Khant, Saket Aryan, Taranjeet Singh, and Deshraj Yadav.
  2025.
\newblock \href {https://doi.org/10.3233/FAIA251160} {{Mem0}: Building
  production-ready {AI} agents with scalable long-term memory}.
\newblock In \emph{ECAI 2025}, volume 413 of \emph{Frontiers in Artificial
  Intelligence and Applications},
  \unskip\nobreakspace{}pages~\mbox{2993--3000}. IOS Press.

\bibitem[{Cook et~al.(2013)Cook, Crandall, Thomas, and
  Krishnan}]{cook2013casas}
Diane~J. Cook, Aaron~S. Crandall, Brian~L. Thomas, and Narayanan~C. Krishnan.
  2013.
\newblock \href {https://doi.org/10.1109/MC.2012.328} {{CASAS}: A smart home in
  a box}.
\newblock \emph{Computer}, \unskip\nobreakspace{}46(7):62--69.

\bibitem[{Grigorik(2026)}]{gharchive}
Ilya Grigorik. 2026.
\newblock \href {https://www.gharchive.org/} {{GH Archive}: A project to record
  the public {GitHub} timeline}.
\newblock Project website.
\newblock \mbox{Accessed August 21, 2026}.

\bibitem[{Guti{\'e}rrez et~al.(2024)Guti{\'e}rrez, Shu, Gu, Yasunaga, and
  Su}]{gutierrez2024hipporag}
Bernal~Jim{\'e}nez Guti{\'e}rrez, Yiheng Shu, Yu~Gu, Michihiro Yasunaga, and
  Yu~Su. 2024.
\newblock \href {https://doi.org/10.52202/079017-1902} {{HippoRAG}:
  Neurobiologically inspired long-term memory for large language models}.
\newblock In \emph{Advances in Neural Information Processing Systems},
  volume~37, \unskip\nobreakspace{}pages~59532--59569.

\bibitem[{Iqbal and Bailey(2007)}]{iqbal2007breakpoints}
Shamsi~T. Iqbal and Brian~P. Bailey. 2007.
\newblock \href {https://doi.org/10.1145/1240624.1240732} {Understanding and
  developing models for detecting and differentiating breakpoints during
  interactive tasks}.
\newblock In \emph{Proceedings of the SIGCHI Conference on Human Factors in
  Computing Systems}, \unskip\nobreakspace{}pages~697--706.

\bibitem[{Iqbal and Bailey(2008)}]{iqbal2008notification}
Shamsi~T. Iqbal and Brian~P. Bailey. 2008.
\newblock \href {https://doi.org/10.1145/1357054.1357070} {Effects of
  intelligent notification management on users and their tasks}.
\newblock In \emph{Proceedings of the SIGCHI Conference on Human Factors in
  \mbox{Computing Systems}}, \unskip\nobreakspace{}pages~\mbox{93--102}.

\bibitem[{Klasnja et~al.(2019)Klasnja, Smith, Seewald, Lee, Hall, Luers,
  Hekler, and Murphy}]{klasnja2019heartsteps}
Predrag Klasnja, Shawna Smith, Nicholas~J. Seewald, Andy Lee, Kelly Hall, Brook
  Luers, Eric~B. Hekler, and Susan~A. Murphy. 2019.
\newblock \href {https://doi.org/10.1093/abm/kay067} {Efficacy of contextually
  tailored suggestions for physical activity: A micro-randomized optimization
  trial of {HeartSteps}}.
\newblock \emph{Annals of Behavioral Medicine},
  \unskip\nobreakspace{}53(6):573--582.

\bibitem[{Li et~al.(2026)Li, Fan, Li, and Que}]{li2026memcog}
Zihan Li, Xingyu Fan, Feifei Li, and Wenhui Que. 2026.
\newblock \href {https://arxiv.org/abs/2605.28046} {{MemCog}: From
  memory-as-tool to memory-as-cognition in conversational agents}.
\newblock \emph{\mbox{arXiv preprint arXiv:2605.28046}}.

\bibitem[{Liu et~al.(2024)Liu, Lin, Hewitt, Paranjape, Bevilacqua, Petroni, and
  Liang}]{liu2024lost}
Nelson~F. Liu, Kevin Lin, John Hewitt, Ashwin Paranjape, Michele Bevilacqua,
  Fabio Petroni, and Percy Liang. 2024.
\newblock \href {https://doi.org/10.1162/tacl_a_00638} {Lost in the middle: How
  language models use long contexts}.
\newblock \emph{Transactions of the Association for Computational Linguistics},
  \unskip\nobreakspace{}12:157--173.

\bibitem[{Lu et~al.(2025)Lu, Yang, Qian, Chen, Luo, Wu, Wang, Cong, Zhang, Lin,
  Liu, Wang, Liu, Liu, and Sun}]{lu2025proactiveagent}
Yaxi Lu, Shenzhi Yang, Cheng Qian, Guirong Chen, Qinyu Luo, Yesai Wu, Huadong
  Wang, Xin Cong, Zhong Zhang, Yankai Lin, Weiwen Liu, Yasheng Wang, Zhiyuan
  Liu, Fangming Liu, and Maosong Sun. 2025.
\newblock \href {https://openreview.net/forum?id=sRIU6k2TcU} {Proactive agent:
  Shifting {LLM} agents from reactive responses to active assistance}.
\newblock In \emph{The Thirteenth International \mbox{Conference on Learning
  Representations}}.

\bibitem[{Maharana et~al.(2024)Maharana, Lee, Tulyakov, Bansal, Barbieri, and
  Fang}]{maharana2024locomo}
Adyasha Maharana, Dong-Ho Lee, Sergey Tulyakov, Mohit Bansal, Francesco
  Barbieri, and Yuwei Fang. 2024.
\newblock \href {https://doi.org/10.18653/v1/2024.acl-long.747} {Evaluating
  very long-term conversational memory of {LLM} agents}.
\newblock In \emph{Proceedings of the 62nd Annual Meeting of the Association
  for Computational Linguistics (Volume 1: Long Papers)},
  \unskip\nobreakspace{}pages~\mbox{13851--13870}.

\bibitem[{Mehrotra et~al.(2016)Mehrotra, Pejovic, Vermeulen, Hendley, and
  Musolesi}]{mehrotra2016myphone}
Abhinav Mehrotra, Veljko Pejovic, Jo~Vermeulen, Robert~J. Hendley, and Mirco
  Musolesi. 2016.
\newblock \href {https://doi.org/10.1145/2858036.2858566} {My phone and me:
  Understanding people's receptivity to mobile notifications}.
\newblock In \emph{Proceedings of the 2016 CHI Conference on Human Factors in
  Computing Systems}, \unskip\nobreakspace{}pages~1021--1032.

\bibitem[{{Mem0}(2026)}]{mem0ai2026mem0v2012}
{Mem0}. 2026.
\newblock \href {https://github.com/mem0ai/mem0/releases/tag/v2.0.12} {{Mem0}
  v2.0.12}.
\newblock Software release.
\newblock Commit 42cf18c4e6adb448e981aa1c7b55c1602b0cb670; released July 13,
  2026; accessed July 18, 2026.

\bibitem[{Packer et~al.(2023)Packer, Wooders, Lin, Fang, Patil, Stoica, and
  Gonzalez}]{packer2023memgpt}
Charles Packer, Sarah Wooders, Kevin Lin, Vivian Fang, Shishir~G. Patil, Ion
  Stoica, and Joseph~E. Gonzalez. 2023.
\newblock \href {https://arxiv.org/abs/2310.08560} {{MemGPT}: Towards {LLM}s as
  operating systems}.
\newblock \emph{arXiv preprint arXiv:2310.08560}.

\bibitem[{Park et~al.(2023)Park, O'Brien, Cai, Morris, Liang, and
  Bernstein}]{park2023generative}
Joon~Sung Park, Joseph~C. O'Brien, Carrie~J. Cai, Meredith~Ringel Morris, Percy
  Liang, and Michael~S. Bernstein. 2023.
\newblock \href {https://doi.org/10.1145/3586183.3606763} {Generative agents:
  Interactive simulacra of human behavior}.
\newblock In \emph{Proceedings of the 36th Annual ACM Symposium on User
  \mbox{Interface Software and Technology}}.

\bibitem[{{THUNLP}(2026)}]{thunlp2026proactiveagentrepo}
{THUNLP}. 2026.
\newblock \href
  {https://github.com/thunlp/ProactiveAgent/tree/3fcf9beebe256b86871659fbb12541c41c9381b9}
  {{ProactiveAgent}: Official repository}.
\newblock GitHub repository.
\newblock Accessed July 18, 2026; commit
  3fcf9beebe256b86871659fbb12541c41c9381b9.

\bibitem[{Wu et~al.(2025)Wu, Wang, Yu, Zhang, Chang, and
  Yu}]{wu2025longmemeval}
Di~Wu, Hongwei Wang, Wenhao Yu, Yuwei Zhang, Kai-Wei Chang, and Dong Yu. 2025.
\newblock \href {https://openreview.net/forum?id=pZiyCaVuti} {{LongMemEval}:
  Benchmarking chat assistants on long-term interactive memory}.
\newblock In \emph{International \mbox{Conference on Learning
  Representations}}.

\bibitem[{Wu et~al.(2026)Wu, Zhang, Zhou, Wang, Peng, Li, Fan, and
  Zhao}]{wu2026remember}
Yifan Wu, Lizhu Zhang, Yuhang Zhou, Mingyi Wang, Bo~Peng, Serena Li, Xiangjun
  Fan, and Zhuokai Zhao. 2026.
\newblock \href {https://arxiv.org/abs/2607.08716} {Remember when it matters:
  Proactive memory agent for long-horizon agents}.
\newblock \emph{\mbox{arXiv preprint arXiv:2607.08716}}.

\bibitem[{Xie et~al.(2026)Xie, Hu, Ye, Zhang, Chai, Liu, Wu, Zhang, Liao, Hu,
  Ye, Miao, and Yan}]{xie2026pask}
Zhifei Xie, Zongzheng Hu, Fangda Ye, Xin Zhang, Haobo Chai, Zihang Liu,
  Pengcheng Wu, Guibin Zhang, Yue Liao, Xiaobin Hu, Deheng Ye, Chunyan Miao,
  and Shuicheng Yan. 2026.
\newblock \href {https://arxiv.org/abs/2604.08000} {{PASK}: Toward intent-aware
  proactive agents with long-term memory}.
\newblock \emph{arXiv preprint arXiv:2604.08000}.
\newblock Technical report; work in progress.

\bibitem[{Xu et~al.(2025)Xu, Liang, Mei, Gao, Tan, and Zhang}]{xu2025amem}
Wujiang Xu, Zujie Liang, Kai Mei, Hang Gao, Juntao Tan, and Yongfeng Zhang.
  2025.
\newblock \href {https://doi.org/10.52202/085713-0593} {{A-Mem}: Agentic memory
  for {LLM} agents}.
\newblock In \emph{Advances in Neural Information Processing Systems},
  volume~38, \unskip\nobreakspace{}pages~\mbox{17577--17604}.

\bibitem[{Yang et~al.(2025)Yang, Xu, Zeng, Liu, Jiang, Lu, Chen, Jiang, Xing,
  and Yan}]{yang2025contextagent}
Bufang Yang, Lilin Xu, Liekang Zeng, Kaiwei Liu, Siyang Jiang, Wenrui Lu,
  Hongkai Chen, Xiaofan Jiang, Guoliang Xing, and Zhenyu Yan. 2025.
\newblock \href {https://doi.org/10.52202/085713-5582} {{ContextAgent}:
  Context-aware proactive {LLM} agents with open-world sensory perceptions}.
\newblock In \emph{Advances in Neural Information Processing Systems},
  volume~38.

\bibitem[{{Zep}(2026)}]{zep2026graphiti}
{Zep}. 2026.
\newblock \href {https://github.com/getzep/graphiti} {{Graphiti}: A framework
  for building temporal knowledge graphs}.
\newblock Open-source software.
\newblock Accessed September 14, 2026.

\bibitem[{Zhong et~al.(2024)Zhong, Guo, Gao, Ye, and
  Wang}]{zhong2024memorybank}
Wanjun Zhong, Lianghong Guo, Qiqi Gao, He~Ye, and Yanlin Wang. 2024.
\newblock \href {https://doi.org/10.1609/aaai.v38i17.29946} {{MemoryBank}:
  Enhancing large language models with long-term memory}.
\newblock \emph{Proceedings of the AAAI Conference on Artificial
  \mbox{Intelligence}}, \unskip\nobreakspace{}38:\mbox{19724--19731}.

\end{thebibliography}
\endgroup
\clearpage
\appendix
\section{Benchmark and annotation protocol}
\label{app:benchmark_protocol}

\subsection{Public events and availability}

\paragraph{Sports.}
HeartSteps~\citep{klasnja2019heartsteps} supplies 1,456 mobile activity meta-events from nine complete participant streams over 367 observed participant-days.
Public fields include decision-time availability, recognized activity, recent steps, location category, weather, and local decision slot.
Services concern optional activity support. Inputs omit future outcomes and randomized treatments.

\paragraph{Home.}
CASAS~\citep{cook2013casas} supplies ambient-sensor traces from four homes over 14 consecutive days each.
Per local day, ten-minute source blocks are assigned exactly once to 28 regular 50-minute segments and one 40-minute tail, yielding 406 events per home.
A segment becomes available only after interval closure and contains within-interval motion counts, room sequences, door actions, and compact numeric summaries.
Gentle check-ins exclude disease, fall, travel, danger, and unsupported preference inference.

\paragraph{Code.}
Public repository activity, including GH Archive records~\citep{gharchive}, supplies 1,459 events across four repositories with 174 reference interventions.
Pushes, issue updates, pull-request actions, and reviews retain GitHub's native event granularity, with one decision per event.
Records are ordered by timestamp and then event ID.
Each repository owns isolated memory and intervention state even when contributors overlap.
Services cover low-risk triage, review, and maintenance; native \texttt{PushEvent} records enter the stream.

\paragraph{Public-input contract.}
Both label classes share the same scope-specific schema: event and stream identifiers, timestamps, and contemporaneously observable fields.
Reference labels, candidate flags, episode annotations, memory-level markers, and scoring masks are reserved for evaluation.
Evaluation includes early events before temporal trajectories or behavioral regularities are available.

\subsection{Reference labels and causal replay}
\label{app:reference_protocol}

Annotators apply the opportunity and novelty semantics in Section~\ref{sec:reference_labels} to each current event and its complete prior history within the same stream, independently of a tested method's memory.
LLM-assisted triage prioritizes the review queue; three team members independently assign labels while blinded to decisions from the evaluated systems, followed by cross-review and adjudication of every public event.

The Home aggregation rule precedes event construction.
Each evaluation version fixes the event inventory, labels, intervention prompt, event memory bound, and validation settings before main runs on the jointly developed benchmark.

Every stream starts from empty method state and is replayed chronologically under the causal execution order in Section~\ref{sec:decision_control}.
Current events are committed only after the output is finalized; maintenance incorporating an event affects later decisions.
Labels are joined offline; reference episode states and scoring metadata are reserved for evaluation.

\subsection{Example events and reference labels}
\label{app:input_event_examples}

Figure~\ref{fig:input_event_examples} shows one public event record from each scope of ProactiveStream.
Identifiers use \texttt{event\_id}; values and nesting are retained, with some fields omitted.
Events are processed chronologically within each stream.
The ground-truth (GT) reference label is shown above each excerpt: $y_t=1$ denotes \emph{intervene} and $y_t=0$ denotes \emph{abstain}.
These labels are stored separately for evaluation and are excluded from agent inputs.

\lstdefinestyle{inputevent}{
  basicstyle=\ttfamily\fontsize{9}{10.5}\selectfont,
  columns=fullflexible,
  keepspaces=true,
  showstringspaces=false,
  upquote=true,
  breaklines=true,
  breakatwhitespace=false,
  breakindent=1em,
  frame=single,
  framerule=0.3pt,
  rulecolor=\color{black!35},
  framesep=3pt,
  xleftmargin=3.5pt,
  xrightmargin=3.5pt,
  aboveskip=5pt,
  belowskip=5pt
}

\begin{figure*}[p]
\centering
\begin{minipage}[t]{0.48\textwidth}
\textbf{Sports.}\par
\textbf{GT:} $y_t=0$ (\emph{abstain}).
\begin{lstlisting}[style=inputevent]
{
  "event_id": "6379c89c927a28f79bc620c3",
  "timestamp": "2015-07-22T21:31:48Z",
  "stream_id": "heartsteps_participant_01",
  "observable_payload": {
    "source_local_decision_time":
      "2015-07-22T17:31:48",
    "source_utc_offset_seconds": -14400,
    "available": true,
    "recognized_activity": "STILL",
    "google_fit_steps_pre30": 0,
    "google_fit_steps_pre60": 73,
    "jawbone_steps_pre30": 261,
    "jawbone_steps_pre60": 456,
    "location_category": "home",
    "weather_condition": "Mostly Cloudy",
    "temperature_c": 25.8
  }
}
\end{lstlisting}
An activity-context record at 17:31:48 local time (UTC$-$04:00), with availability, location, weather, and step counts from Google Fit and Jawbone over the preceding 30 and 60 minutes.
\end{minipage}\hfill
\begin{minipage}[t]{0.48\textwidth}
\textbf{Home.}\par
\textbf{GT:} $y_t=0$ (\emph{abstain}).
\begin{lstlisting}[style=inputevent]
{
  "event_id": "casas50_d_000000",
  "stream_id": "casas_home_0001",
  "window_start": "2012-07-21T00:00:00",
  "window_end": "2012-07-21T00:50:00",
  "decision_time": "2012-07-21T00:50:00",
  "duration_minutes": 50,
  "motion": {
    "raw_event_count": 84,
    "activation_count": 43,
    "deactivation_count": 41,
    "location_counts": {
      "Bathroom": 13,
      "Bedroom": 17,
      "LivingRoom": 11,
      "OutsideDoor": 2
    },
    "location_sequence": [
      "Bedroom", "LivingRoom", "OutsideDoor",
      "Bedroom", "Bathroom", "Bedroom"
    ]
  },
  "door_actions": [],
  "numeric_sensors": [],
  "other_actions": []
}
\end{lstlisting}
This event summarizes sensor observations from 00:00 to 00:50 in source-local time and becomes available when the interval closes at 00:50.
Location counts record sensor activations; the location sequence collapses consecutive repeated location labels.
\end{minipage}

\par\medskip
\begin{minipage}{\textwidth}
\textbf{Code.}\par
\textbf{GT:} $y_t=1$ (\emph{intervene}).
\begin{lstlisting}[style=inputevent]
{
  "event_id": "ghe_51480602428",
  "event_time": "2025-06-30T00:22:02.000Z",
  "stream_id": "github_repo_169435703",
  "repository": {"full_name": "callstack/react-native-pager-view"},
  "event": {"action": "created", "github_type": "IssueCommentEvent"},
  "object": {"number": 1005, "type": "issue"},
  "content": {
    "title": "java.lang.IllegalArgumentException: Scrapped or attached views may not be recycled",
    "comment_body": "In my case, I was able to easily reproduce this by starting to scroll to view the next screen and then going back before the scroll finished."
  },
  "attributes": {"state": "open"}
}
\end{lstlisting}
\textit{Prior context.} On June 23, a collaborator requested reproduction details for this crash, and the reporter remained unable to reproduce it.
\textit{Intervention opportunity.} The June 30 comment supplies a concrete reproduction sequence, making targeted validation of the unresolved crash actionable.
\end{minipage}
\caption{Example input events and GT reference labels from Sports, Home, and Code. Excerpts preserve the original values and nesting; labels are shown separately from the inputs.}
\label{fig:input_event_examples}
\end{figure*}

\section{Implementation and experimental configuration}
\label{app:implementation}

\subsection{Trajectory evidence and replay}

The trajectory construction step uses an independent buffer of all committed events from the open local day, cleared after processing.
Activity segments contain a local interval, typed \texttt{observed}, \texttt{context-before}, and \texttt{outcome} fields, plus source event identifiers.
The checks in Equation~\ref{eq:daily_validation} canonicalize scalar values, accept exact or schema-preserving suffix field paths, and require consistent entities, intervals, and resolved citations.
Validation and replay operate on typed evidence.
Accepted trajectories remain in the current construction cycle after their events leave event memory.

Empty or unobservable records also advance the seven-record construction cycle.
Replay follows Equation~\ref{eq:replay_gate}: a triggered, observable day is fulfilled when the expected behavior occurs at or after the trigger within the window; otherwise it is violated.
Missing observations are censored, each date counts at most once, and the base rate includes triggered and other observable days under the same time and weekday restrictions.
Confidence and lift have zero-denominator scores of zero.

\subsection{Lifecycle and current matching}
\label{app:lifecycle}

After admission, confidence combines replay and online counts in a support ratio smoothed with one initial success and failure; censored observations preserve both counts.
Two consecutive violations move an active regularity to \emph{weakening}, lowering retrieval priority; four move it to \emph{dormant}.
A fulfilled observation restores a weakening regularity to active at confidence $\geq0.5$.
Dormant regularities return to candidate for replay readmission.

\paragraph{Current matching.}
\label{app:matching}
Current matching is deterministic.
Retrieval admits upcoming, pending, or deviated occurrences of active or weakening regularities.
Occurrence priority is deviated, followed by pending and upcoming.

\subsection{Intervention control}
\label{app:intervention_control}

In Sports and Home, the intervention key combines scope-specific episode state, proposed service action and target, and sorted regularity identifiers available in context.
Plain summaries or graph facts supply an empty list of regularity identifiers.
Sports state includes availability and activity; Home state includes activity and room and onset, confirmed, and sustained phases at one, two, and three or more consecutive matching observations.
Without an explicit episode key, scope-specific time or target fields identify the opportunity, with unchanged evidence additionally required for duplicate suppression.
State changes, sequence restarts, changed service or regularity identifiers, and local-day transitions can generate new keys.

Code uses the identity of the repository and object, along with the most recent allowed intervention's mode, recipient, content, and firing condition.
An exact match converts a repeated \texttt{create} proposal to \emph{abstain}; changed signatures and other episode operations pass this check.
Representation-dependent key evidence can yield different suppression decisions across memory interfaces.

\subsection{Model and method interfaces}
\label{app:configuration}

Decision and memory-construction calls use the model, API, and decoding settings in Section~\ref{sec:experimental_settings}, with separate output budgets.
Decision limits are interface-specific (256 or 512 tokens); construction limits are configured by call role.
Table~\ref{tab:baseline_interfaces} consolidates the external-method interfaces.
Each method retains its prompts, retrieval, and budgets.

\begin{table*}[t]
\centering
\caption{External-method memory interfaces and implementation settings; all methods share the backbone.}
\label{tab:baseline_interfaces}
\begin{tabular}{@{}p{0.16\textwidth}p{0.80\textwidth}@{}}
\toprule
\textbf{Method} & \textbf{Interface and Settings} \\
\midrule
ProactiveAgent & Released prompt and proposal/null contract on the study backbone; causal event--decision history capped at 1,000 events per request including the current event~\citep{lu2025proactiveagent,thunlp2026proactiveagentrepo}. \\
\addlinespace[3pt]
ContextAgent & In-context proactive-service judgment on public events~\citep{yang2025contextagent}. \\
\addlinespace[3pt]
Mem0 & Open-source \texttt{mem0ai==2.0.12} extraction, consolidation, and retrieval; local \texttt{all-MiniLM-L6-v2} embeddings; retrieval of up to eight memories for the same entity with search threshold $0.1$~\citep{chhikara2025mem0,mem0ai2026mem0v2012}. \\
\addlinespace[3pt]
A-MEM & Linked notes with association updates augment bounded context~\citep{xu2025amem}. \\
\addlinespace[3pt]
Graphiti & Graphiti 0.30.2 with Neo4j and local 384-dimensional \texttt{all-MiniLM-L6-v2} embeddings; committed events populate separate graphs for each stream. Queries from the current event use native hybrid edge search with reciprocal-rank fusion, retrieving up to ten relation facts with validity fields and source IDs~\citep{zep2026graphiti}. \\
\bottomrule
\end{tabular}
\end{table*}

Graphiti supplies relation facts directly, using the task policy and output contract with memory-specific guidance; its ingestion episodes correspond to source records.

\paragraph{Scoring.}
Using final-decision confusion counts, relevance is $\tfrac{\mathrm{TP}}{\mathrm{TP}+\mathrm{FP}}$, coverage is $\tfrac{\mathrm{TP}}{\mathrm{TP}+\mathrm{FN}}$, effectiveness is $\tfrac{2\mathrm{TP}}{2\mathrm{TP}+\mathrm{FP}+\mathrm{FN}}$, and intrusion is $\tfrac{\mathrm{FP}}{\mathrm{FP}+\mathrm{TN}}$.
Balance is $\tfrac{1}{2}[\tfrac{\mathrm{TP}}{\mathrm{TP}+\mathrm{FN}}+\tfrac{\mathrm{TN}}{\mathrm{TN}+\mathrm{FP}}]$.
For frequency, $1\times$ indicates equal counts of interventions and reference-positive events; larger and smaller values indicate more and fewer interventions, respectively.
Frequency is interpreted jointly with relevance, coverage, and intrusion.

\subsection{Cross-family event-only configuration}
\label{app:cross_family_streams}

Table~\ref{tab:cross_family_streams} lists the two complete streams selected within each scope.
Streams are ranked by SHA256 of \texttt{seed|}\allowbreak\texttt{domain|}\allowbreak\texttt{public\_stream\_id}.
The fixed seed is \texttt{proactivestream-}\allowbreak\texttt{event-only-}\allowbreak\texttt{nine-streams-}\allowbreak\texttt{20260925-v1}; the two lowest hashes in each scope determine selection.
The selection uses public stream identifiers.

\begin{table}[t]
\centering
\caption{Streams in the cross-family comparison.}
\label{tab:cross_family_streams}
\begin{tabular}{@{}llr@{}}
\toprule
\textbf{Scope} & \textbf{Source Stream} & \textbf{Events} \\
\midrule
Sports & HeartSteps participant 05 & 184 \\
 & HeartSteps participant 35 & 191 \\
Home & CASAS home 0004 & 406 \\
 & CASAS home 0003 & 406 \\
Code & \texttt{Unitech/pm2} & 365 \\
 & \texttt{thephpleague/commonmark} & 347 \\
\midrule
Total & six complete streams & 1,899 \\
\bottomrule
\end{tabular}
\end{table}

Each stream is replayed chronologically from empty state.
Models share the scope-specific task prompt, intervention control policy, and a causal context containing the current event and up to 99 prior events from the same stream within 72 hours.
Requests use temperature zero and request disabled thinking, with decision output limits of 256 tokens for Sports and Home and 512 for Code.
Output-format and transport-recovery settings follow each backend's interface.
Final decisions are scored offline against the common reference labels.

\section{Additional ablation and intervention control results}
\label{app:extended_ablations}

\looseness=-1
Table~\ref{tab:validation_ablations_full} supplements the relevance and effectiveness scores in Table~\ref{tab:validation_ablations} with frequency and coverage.
On Code, suppression removes 66 false-positive and 17 true-positive proposals; coverage falls from 99.4\% to 89.7\%.
The diagnostic scores recorded proposals while holding later states fixed.
In Figure~\ref{fig:memory_hierarchy}, the event and trajectory memory variant records 4, 15, and 64 more true positives than event-only on Sports, Home, and Code, respectively; \system{} records a further 57, 33, and 22.

\begin{table}[t]
\centering
\caption{Additional validation and Code intervention control scores. Frequency uses multiples ($\times$); coverage uses percentages. Bold and underline mark the best and second-best coverage per scope, including ties.}
\label{tab:validation_ablations_full}
\resizebox{\columnwidth}{!}{%
\begin{tabular}{@{}clcc@{}}
\toprule
\textbf{Scope} & \textbf{Method} & \begin{tabular}[c]{@{}c@{}}\textbf{Frequency}\\ ($\times$)\end{tabular} & \begin{tabular}[c]{@{}c@{}}\textbf{Coverage}\\ $\uparrow$\end{tabular} \\
\midrule
\multirow{3}{*}{Sports} & w/o replay gate & 2.19 & \underline{68.2} \\
 & w/o trajectory validation & 2.23 & \textbf{76.7} \\
 & \system{} & 2.15 & \textbf{76.7} \\
\midrule
\multirow{3}{*}{Home} & w/o replay gate & 1.65 & 72.1 \\
 & w/o trajectory validation & 1.62 & \textbf{74.2} \\
 & \system{} & 1.52 & \underline{73.7} \\
\midrule
\multirow{4}{*}{Code} & w/o replay gate & 2.47 & 83.3 \\
 & w/o trajectory validation & 2.48 & \underline{89.7} \\
 & before suppression & 2.71 & \textbf{99.4} \\
 & \system{} & 2.24 & \underline{89.7} \\
\bottomrule
\end{tabular}%
}
\end{table}

\section{Reference and episode diagnostics}
\label{app:label_review_450}

\begin{table*}[!t]
\centering
\caption{Reference sensitivity on 150 targets per scope. Effectiveness (\%) scores identical outputs under earlier and separate-team references. Positives and confusion counts use the latter; $I$ counts final interventions.}
\label{tab:reference_sensitivity_450}
\begin{tabular}{lrrrrrrcc}
\toprule
\textbf{Scope} & \textbf{Positives} & $\boldsymbol{I}$ & \textbf{TP} & \textbf{FP} & \textbf{FN} & \textbf{TN} & \textbf{\shortstack{Earlier Diagnostic\\Effectiveness}} & \textbf{\shortstack{Separate-Team\\Effectiveness}} \\
\midrule
Sports & 27 & 86 & 24 & 62 & 3 & 61 & 42.6 & 42.5 \\
Home & 15 & 39 & 15 & 24 & 0 & 111 & 68.7 & 55.6 \\
Code & 34 & 42 & 18 & 24 & 16 & 92 & 57.6 & 47.4 \\
\bottomrule
\end{tabular}
\end{table*}

\begin{table*}[!t]
\centering
\caption{\system{} episode diagnostic on 150 targets per scope. Coverage, SR, and RIR use percentages. Unresolved denotes ambiguous semantic or condition matches; uncertain follow-ups ($U$) are zero in all scopes.}
\label{tab:episode_review_450}
\begin{tabular}{lcrrrrrrrcc}
\toprule
\textbf{Scope} & \textbf{\shortstack{Covered/New\\(Coverage)}} & $\boldsymbol{I}$ & $\boldsymbol{F}$ & $\boldsymbol{V}$ & $\boldsymbol{R}$ & \textbf{Unmatched} & \textbf{Late} & \textbf{Unresolved} & \textbf{SR} & \textbf{RIR} \\
\midrule
Sports & 23/26 (88.5) & 86 & 23 & 1 & 1 & 60 & 0 & 1 & 27.9 & 1.2 \\
Home & 9/9 (100.0) & 39 & 10 & 6 & 17 & 6 & 0 & 0 & 41.0 & 43.6 \\
Code & 14/31 (45.2) & 42 & 14 & 3 & 4 & 16 & 4 & 1 & 40.5 & 9.5 \\
\bottomrule
\end{tabular}
\end{table*}

\subsection{Reference sensitivity}

The diagnostic selects 150 consecutive targets from one stream per scope using a fixed SHA256 ranking of eligible streams and contiguous starts based only on public metadata, with at least seven days of preceding history.
The streams are Sports participant\_35, Home home\_0004, and Code repository \texttt{callstack/react-native-pager-view}.
Their target periods are December 16, 2015--January 18, 2016; June 23--28, 2011; and June 21--November 4, 2025.
All 37, 203, and 152 preceding events are processed chronologically, respectively: 842 public events supply context for 450 scored targets.

A separate team of three reviewers independently labels each target using prior history from the same stream, blinded to outputs from the evaluated systems, then discusses judgments to obtain consensus references.
Earlier references use original Sports and Home labels and 122 original plus 28 additional Code annotations.
Relative to these references, Sports and Home each have 13 changes from \emph{intervene} to \emph{abstain}, with 4 and 0 changes in the reverse direction, respectively; Code has 6 removals and 16 additions to positive labels.

Table~\ref{tab:reference_sensitivity_450} scores identical recorded \system{} outputs against these two sets of references.
The predictions come from separate diagnostic runs on the selected targets.
The diagnostic uses Luna with temperature zero, thinking disabled, decision limits of 256 tokens for Sports and Home and 512 for Code, and the same context bound of 72 hours and at most 100 events, trajectory validation, replay gating, and intervention control policy.
Immediate and scheduled services both count as \emph{intervene}.

Rescoring against the separate-team references yields similar effectiveness on Sports and lower effectiveness on Home and Code relative to the earlier references.
Absolute scores are label-sensitive, especially on Home and Code.

\subsection{Episode references and matching}
\label{app:episode_review_450}

A model-assisted post-hoc analysis on these targets reconstructs episodes defined by service goals and objects from positive event anchors.
Each reference specifies a service goal and object, causal evidence, onset, explicitly listed valid observations, and any deadline or closure evidence.
Semantic matching uses these fields to associate interventions with reference episodes.

\paragraph{Matching and follow-up.}
Episode coverage follows \system{}'s own interventions, initialized from recorded prefix outputs.
Only target interventions are scored.
An intervention must match the goal and object at a listed valid observation.
Its proposed service time is the decision time plus its specified delay (zero for immediate service), and must meet any explicit episode deadline.
The first timely match covers the episode.
Each output receives at most one credit: the first uncovered compatible episode in the fixed reference order, otherwise the first already covered one.

Subsequent matches are compared with the method's preceding assistance for that episode.
A valuable follow-up requires new causal evidence that changes the actionable service or next step, and a message addressing that change.
Rephrasing or continued status alone is redundant; value requiring unobserved delivery, recipient response, or completion remains uncertain.
Each reference episode receives coverage credit once.

\paragraph{Metrics and boundaries.}
Let $I$ count all target interventions, $F$ first valid interventions across all reference episodes, $V$ valuable follow-ups, $R$ redundant follow-ups, and $U$ uncertain follow-ups.
New reference episodes have onsets within the target interval.
Episode coverage divides covered new episodes by all new reference episodes; supported relevance is $\mathrm{SR}=(F+V)/I$, and redundant intervention rate is $\mathrm{RIR}=R/I$.
Uncertain, unmatched, late, and unresolved outcomes contribute only to the intervention count $I$.
Table~\ref{tab:episode_review_450} combines the rates and outcome counts.

Episode coverage is highest on Home and lowest on Code; the redundant-intervention share is highest on Home and lowest on Sports.

The exploratory scores depend on reference construction and semantic matching.

\FloatBarrier

\end{document}